\pdfoutput=1
\documentclass[11pt]{article}
\usepackage{style/emnlp2021}
\usepackage{times}
\usepackage{latexsym}
\usepackage[T1]{fontenc}
\usepackage[utf8]{inputenc}
\usepackage{microtype}
\usepackage{soul}
\usepackage[T1]{fontenc}

\newcommand{\figfile}[1]{2021_emnlp_answer_equiv/figures/#1}

\title{What's in a Name?  \\ Answer Equivalence For Open-Domain Question Answering}

\author{
  Chenglei Si \\
 Computer Science \\
University of Maryland \\
\tt{clsi@terpmail.umd.edu} \\
\And
Chen Zhao \\
Computer Science \\
University of Maryland \\ 
\tt{chenz@cs.umd.edu} \\
\And
Jordan Boyd-Graber \\
\abr{cs}, \abr{lsc}, \abr{umiacs}, and iSchool \\
University of Maryland \\
\tt{jbg@umiacs.umd.edu}
}

\date{}

\newif\ifcomment\commenttrue

\usepackage[a-1b]{pdfx}

\usepackage{framed}
\usepackage{mdwlist}
\usepackage{siunitx}
\usepackage{latexsym}
\usepackage{colortbl}
\usepackage{xcolor}
\usepackage{nicefrac}
\usepackage{booktabs}
\usepackage{fnpct}
\usepackage{amsfonts}
\usepackage[T1]{fontenc}
\usepackage{bold-extra}
\usepackage{amsmath}
\usepackage{amssymb}
\usepackage{bm}
\usepackage{graphicx}
\usepackage{mathtools}
\usepackage{microtype}
\usepackage{multirow}
\usepackage{multicol}
\usepackage{xpatch}
\usepackage{latexsym,comment}
\usepackage[normalem]{ulem}

\newcommand*{\missingreference}{{\Huge \colorbox{red}{?reference?}}}
\newcommand*{\missingcitation}{{\Huge \colorbox{red}{?citation?}}}

\makeatletter
\xpatchcmd{\@setref}{\bfseries}{\missingreference}{}{}
\def\@citex[#1]#2{\leavevmode
    \let\@citea\@empty
    \@cite{\@for\@citeb:=#2\do
        {\@citea\def\@citea{,\penalty\@m\ }%
            \edef\@citeb{\expandafter\@firstofone\@citeb\@empty}%
            \if@filesw\immediate\write\@auxout{\string\citation{\@citeb}}\fi
            \@ifundefined{b@\@citeb}{\hbox{\reset@font\missingcitation}%
                \G@refundefinedtrue
                \@latex@warning
                {Citation `\@citeb' on page \thepage \space undefined}}%
            {\@cite@ofmt{\csname b@\@citeb\endcsname}}}}{#1}}
\makeatother

\newcommand{\gem}[1]{\mbox{\textsc{gem}}}
\newcommand{\abr}[1]{\textsc{#1}}
\newcommand{\camelabr}[2]{{\small #1}{\textsc{#2}}}

\newcommand{\hidetext}[1]{}
\newcommand{\ignore}[1]{}

\ifcomment
    \newcommand{\pinaforecomment}[3]{\colorbox{#1}{\parbox{.8\linewidth}{#2: #3}}}

    \newcommand{\prtodo}[1]{\pinaforecomment{lightblue}{pr}{#1}}
    \newcommand{\prtodoi}[1]{\pinaforecomment{lightblue}{pr}{#1}}
\else
    \newcommand{\pinaforecomment}[3]{}
    \newcommand{\prtodo}[1]{}
    \newcommand{\prtodoi}[1]{}
\fi

\newcommand{\smallurl}[1]{ \begin{tiny}\url{#1}\end{tiny}}

\definecolor{lightblue}{HTML}{3cc7ea}
\definecolor{CUgold}{HTML}{CFB87C}
\definecolor{grey}{rgb}{0.95,0.95,0.95}
\definecolor{ceil}{rgb}{0.57, 0.63, 0.81}
\definecolor{UMDred}{HTML}{ed1c24}
\definecolor{UMDyellow}{HTML}{ffc20e}

\newcommand{\qa}[0]{\abr{qa}}
\newcommand{\odqa}[0]{\abr{odqa}}

\newcommand{\triviaqa}{\camelabr{Trivia}{qa}}
\newcommand{\ambigqa}{\camelabr{Ambig}{qa}}

\newcommand{\squad}{\textsc{sq}{\small u}\textsc{ad}}
\newcommand{\nq}[0]{\abr{nq}}

\newcommand{\hotpot}{\abr{h}{\small otpot}\abr{qa}}

\newcommand{\bert}{\abr{bert}}

\begin{document}
\maketitle

\begin{abstract}
  
A flaw in \abr{qa} evaluation is that annotations often only provide
one gold answer.
Thus, model predictions semantically equivalent to the answer but
superficially different are considered incorrect.
%
%
This work explores mining alias entities from knowledge bases and using
them as additional gold answers (i.e., equivalent answers).
We incorporate answers for two settings: evaluation with additional
answers and model training with equivalent answers.
We analyse three  \abr{qa} benchmarks: Natural
Questions, \triviaqa{} and \squad{}.
Answer expansion increases the exact match score on all datasets for evaluation, while incorporating it helps model
training over real-world datasets.
We ensure the additional answers are valid through a
human \textit{post hoc} evaluation.\footnote{Our code and data are available at: \url{https://github.com/NoviScl/AnswerEquiv}.}

\end{abstract}

\section{Introduction: A Name that is the Enemy of Accuracy}
\label{sec:intro}


In question answering (\abr{qa}), computers---given a
question---provide the correct answer to the question.
However, the modern formulation of \abr{qa} usually assumes that each
question has only one answer, \textit{e.g.},
\squad{}~\cite{SQuAD}, 
\hotpot{}~\cite{HotpotQA}, \abr{drop}~\cite{DROP}.
This is often a byproduct of the prevailing framework for modern
\abr{qa}~\cite{DrQA,DPR}: a retriever finds passages that may contain
the answer, and then a machine reader identifies \emph{the} (as in
only) answer span.

In a recent position paper, \citet{QA-TriviaNerds} argue that this is
at odds with the best practices for human \abr{qa}.
This is also a problem for computer \abr{qa}.
%
%
\begin{figure}[t]
  \centering
  \includegraphics[width=0.99\linewidth]{\figfile{ex_fig.pdf}}
  \caption{An example from \abr{nq} dataset. The correct model (\abr{bert}) prediction does not match
  the gold answer but matches the equivalent answer we mine from a knowledge base.}
  \label{fig:ex}
  \end{figure}
  A \bert{} model~\cite{BERT} trained on Natural
  Questions~\cite[\abr{nq}]{NQ} answers \uline{Tim Cook} to
  the question ``Who is the Chief Executive Officer of Apple?''
  (Figure~\ref{fig:ex}), while the gold answer is only \uline{Timothy
    Donald Cook}, rendering \uline{Tim Cook} as wrong as \uline{Tim
    Apple}.
In the 2020 NeurIPS Efficient \abr{qa} competition~\cite{EfficientQA},
human annotators rate nearly a third of the predictions that do not
match the gold annotation as ``definitely
correct'' or ``possibly correct''.

Despite the near-universal acknowledgement of this problem, there is
neither a clear measurement of its magnitude nor a consistent best
practice solution.
While some datasets provide comprehensive answer
sets~\cite[\textit{e.g.},][]{TriviaQA}, subsequent datasets such as \abr{nq} have
not\dots and we do not know whether this is a problem.
We fill that lacuna.


%
Section~\ref{sec:method} mines knowledge bases for alternative
answers to named entities.
Even this straightforward approach finds high-precision answers
not included in official answer sets.
We then incorporate this in both \emph{training} and \emph{evaluation}
of \abr{qa} models to accept alternate answers.
We focus on three popular open-domain \abr{qa} datasets:
\abr{nq}, \triviaqa{} and \squad{}.
Evaluating models with a more permissive evaluation improves exact
match (\abr{em}) by 4.8 points on \triviaqa{}, 1.5 points on \abr{nq},
and 0.7 points on \squad{} (Section~\ref{sec:results}).
By augmenting training data with answer sets, state-of-the-art models 
improve on \abr{nq} and \triviaqa{}, but not on \squad{}
(Section~\ref{sec:analysis}), which was created with a single evidence
passage in mind.
In constrast, augmenting the answer allows diverse evidence sources 
to provide an answer.
After reviewing other approaches for incorporating ambiguity in
answers (Section~\ref{sec:related}), we discuss how to further make
\abr{qa} more robust.


\section{Method: An Entity by any Other Name}
\label{sec:method}

This section reviews the open-domain \abr{qa} (\odqa{}) pipeline and
 introduces how we expand gold answer sets for both training
and evaulation.
 
\subsection{\odqa{} with Single Gold Answer}



We follow the state-of-the-art retriever--reader pipeline for
\odqa{}, where a retriever finds a handful of
passages from a large corpus (usually Wikipedia), then a reader,
often multi-tasked with passage reranking, selects a span as the
prediction.

We adopt a dense passage retriver~\cite[\abr{dpr}]{DPR} to find passages.
\abr{dpr} encodes questions and passages
into dense vectors.
\abr{dpr} searches for passages in this dense space: given an encoded
query, it finds the nearest passage vectors in the dense space.
We do not train a new retriever but instead use the released \abr{dpr} checkpoint
to query the top-$k$ (in this paper, $k=100$) most relevant passages.






Given a question and retrieved passages, a neural \emph{reader} reranks the
top-$k$ passages and extracts an answer span.
Specifically, \abr{bert} encodes each passage~$p_i$ concatenated
with the question~$q$
as~$P_i^{L \times h} = \mbox{\abr{bert}}([p_i;q])$, where $L$
is the maximum sequence length and~$h$ is the hidden size
of \abr{bert}.
Three probabilities use this representation to reveal where we can find the answer.
The first probability~$P_{r}(p_i)$ encodes whether passage~$i$ contains the
answer.
Because the answer is a subset of the longer span, we must provide the
index where the answer starts~$j$ and where it ends~$k$.
Given the encoding of passage~$i$, there are three parameter
matrices~$\textbf{w}$ that produce these probabilities:
\begin{align}
P_{r}(p_i) = & \mbox{softmax}( \textbf{w}_{r}(\textbf{P}_i[0, :])
               ); \label{eq:passage} \\
P_{s}(t_j) = & \mbox{softmax}( \textbf{w}_{s}(\textbf{P}_i[j,:]) ); \label{eq:spanstart} \mbox{and}\\
P_{e}(t_k) = & \mbox{softmax}( \textbf{w}_{e}(\textbf{P}_i[k,:]) ). \label{eq:spanend}
\end{align}
where $\textbf{P}_i[0,:]$ represents the \texttt{[CLS]} token, and
$\textbf{w}_{r}, \textbf{w}_{s}$ and $\textbf{w}_{e}$ are learnable
weights for passage selection, start span and end span.
Training updates weights with one positive and $m-1$ negative passages among the top-100 retrieved passages for each question (we use $m=24$) with 
log-likelihood of the positive passage for passage selection (Equation~\ref{eq:passage}) and
maximum marginal likelihood over all spans in the positive passage for
span extraction (Equations~\ref{eq:spanstart}--\ref{eq:spanend}).


To study the effect of equivalent answers in reader training, we focus
on the distant supervision setting where we know \emph{what} the answer is
but not \emph{where} it is (in contrast to full supervision where we
know both).
To use the answer to discover positive passages, we use string
matching: any of the top-$k$ retrieved passages
that contains an answer is considered correct.
We discard questions without any positive passages.
This framework is consistent with modern state-of-the-art \abr{odqa}
pipelines~\cite[\emph{inter
alia}]{NQ-baseline,DPR,Transformer-XH,GraphRetriever,zhao2021beamdr}.
%
%

\subsection{Extracting Alias Entities}


We expand the original gold answer set by extracting aliases
from Freebase~\cite{Freebase}, a large-scale knowledge base
(\abr{kb}).
Specifically, for each answer in the
original dataset (\textit{e.g.}, \ul{Sun Life Stadium}), if we can find this
entry in the \abr{kb}, we then use the ``common.topic.alias'' 
relation to extract all aliases of the entity (\textit{e.g.}, \ul{[Joe Robbie
  Stadium}, \ul{Pro Player Park}, \ul{Pro Player Stadium}, \ul{Dolphins Stadium}, \ul{Land
  Shark Stadium}]).
We expand the answer set by adding all
aliases.
We next describe how this changes evaluation and training.


\subsection{Augmented Evaluation}



For evaluation, we report the exact match (\abr{em}) score, 
where a predicted span is correct only if the (normalized) span text matches 
with a gold answer exactly.
This is the adopted metric for span-extraction datasets in most
\abr{qa} papers~\cite[\emph{inter
alia}]{DPR,ORQA,DiscreteEM}.
When we incorporate the alias entities in evaluation, we get an
expanded answer set $\mathcal{A} \equiv \{a_1, ..., a_n\}$.
For a given span $s$ predicted by
the model, we compute \abr{em} score of $s$ if the span matches \emph{any} 
correct answer~$a$ in the set $\mathcal{A}$:
\begin{equation}
\mbox{\abr{em}}(s, \mathcal{A}) = \max_{a \in \mathcal{A}}
\{ \mbox{\textsc{em}}(s, a) \}.
\end{equation}


\subsection{Augmented Training}



When we incorporate the alias entities in training, 
we treat each retrieved passage as positive if it contains either the original answer
or the extracted alias entities.
%
As a result, some originally negative passages 
become positive since they may contain the aliases, and we
augment the original training set.
Then, we train on this augmented training set in the same way as in 
Equations~\ref{eq:passage}--\ref{eq:spanend}.

\section{Experiment: Just as Sweet}
\label{sec:results}



\begin{table}[t]
  \small
  \footnotesize
  \center
  \begin{tabular}{lp{0.9cm}p{0.8cm}p{1cm}}
  \toprule
   & {\bf \abr{nq}} & {\bf \squad{}} & {\bf \triviaqa{}}  \\
   \midrule
  Avg. Original Answers & 1.74 & 1.00  & 1.00 \\
  Matched Answers (\%) & 71.63 & 32.16 & 88.04 \\
  Avg. Augmented Answers & 13.04  & 5.60 & 14.03 \\
  \midrule
  \#Original Positives &  69205 & 48135  & 62707 \\
  \#Augmented Positives & 69989  & 48615 & 67526 \\
   \bottomrule
  \end{tabular}
  \caption{\textbf{Avg. Original Answers} denotes the average number of answers
    per question in the official test sets.  \textbf{Matched Ans.}
    denotes the percentage of original answers that have aliases
     in the \abr{kb}.  \textbf{Avg. Augmented Answers} denotes the average number
    of answers in our augmented answer sets.  \textbf{Last two rows}:
    number of positive questions (questions with matched positive
    passages) in the original / augmented training set for each
    dataset.  \abr{nq} and \triviaqa{} have more augmented answers than \squad{}.}


  \label{tab:train_test_stats}
  \end{table}

\begin{table}[t]
    \small
    \center
    \begin{tabular}{llcc}
    \toprule
    {\bf Data}&{\bf Model} & {\bf Single Ans} &  {\bf Ans Set}   \\
     \midrule
     \nq{} 
     &Baseline & 34.9 & 36.4 \\
     &+ Augment Train & 35.8 & 37.2 \\
      \rowcolor{gray!50}
      \abr{TriviaQA} & Baseline & 49.9 & 54.7 \\
      \rowcolor{gray!50}
     &+ Augment Train  & 50.0 & 55.9 \\
\abr{SQuAD}&Baseline & 18.9 & 19.6 \\
     &+ Augment Train & 18.3 &18.9\\
     \bottomrule
    \end{tabular}
    \caption{
    Evaluation results on \abr{qa} datsets compared to the original ``Single
    Ans'' evaluation under the original answer set, using the augmented
    answer sets (``Ans Set'') improves evaluation.  
    Retraining the reader with augmented answer sets (``Augment Train'') 
    is even better for most datasets, even when evaluated on
    the datasets' original answer sets.
    Results are the average of three random seeds.
  }
    \label{tab:main_results}
\end{table}

\begin{table}[t]
\small
\center
\begin{tabular}{lccc}
\toprule
& Baseline & +Wiki Train & +FB Train \\
\midrule
Single Ans. & 49.31 & 49.42  & 49.53 \\
+Wiki Eval & 54.13 & 55.27  & 54.57  \\
+FB Eval & 51.75 & 52.23  & 52.52 \\
\bottomrule
\end{tabular}
\caption{Results on TriviaQA. Numbers in brackets indicate the improvement compared to the first column. Each column indicates a different training setup and each row indicates a different evaluation setup. Augmented training with Wikipedia aliases (2nd column) and Freebase aliases (3rd column) improve EM over baseline (1st column).}
\label{tab:trivia_results}
\end{table}

We present results on three \abr{qa} datasets---\nq{},
\triviaqa{} and \squad{}---on how including aliases as
alternative answers impacts \textit{evaluation} and \textit{training}.
Since the official test sets are not released, we use the original dev sets as the test sets, and randomly split 10\% training data as the held-out dev sets. 
All of these datasets are extractive \abr{qa} datasets where answers are spans in Wikipedia articles. 
%
%
%


\paragraph{Statistics of Augmentation.}
Both \squad{} and \triviaqa{} have one single answer in the test set
(Table~\ref{tab:train_test_stats}).
While \abr{nq} also has answer sets, these represent
annotator ambiguity given a passage, not the full range of possible
answers.
For example, different annotators might variously highlight
\uline{Lenin} or \uline{Chairman Lenin}, but there is no expectation
to exhaustively enumerate all of his names (e.g., \uline{Vladimir
  Ilyich Ulyanov} or \uline{Vladimir Lenin}).
Although the default test set of \triviaqa{} uses one single gold
answer, the authors released answer aliases minded from
Wikipedia. 
Thus, we directly use those aliases for our experiments in Table~\ref{tab:main_results}.
Overall, a systematic approach to expand gold answers significantly increases
gold answer numbers.

\triviaqa{} has the most answers that have equivalent answers, while
\squad{} has the least.
%
%
Augmenting the gold answer set increases the positive passages and thus
%
increases the training examples, since questions with no positive passages are discarded 
(Table~\ref{tab:train_test_stats}), particularly for \triviaqa{}'s
entity-centric questions.

\paragraph{Implementation Details.} 
For all experiments, we use the \texttt{multiset.bert-base-encoder} checkpoint of DPR as the retriever
and use \texttt{bert-base-uncased} for our reader model. During training, we sample one positive 
passage and 23 negative passages for each question. During evaluation, we consider the top-10 retrieved passages for answer span extraction. We use batch size of 16 and learning rate of 3e-5
for training on all datasets.

\paragraph{Augmented Evaluation.}
We train models with the original gold answer set and evaluate under
two settings: 1) on the original gold answer test set; 2) on the
answer test set augmented with alias entities.
%
%
%
On all three datasets, \abr{em} score improves
(Table~\ref{tab:main_results}).
\triviaqa{} shows the largest improvement, as most answers in
\triviaqa{} are entities ($93\%$).

\paragraph{Augmented Training.}
We incorporate the alias answers in training and compare the
results with single-answer training
(Table~\ref{tab:main_results}).
One check that this is encouraging the models to be more robust and
not a more permissive evaluation is that augmented
training improves \abr{em} by about a point even {\bf on the original
  single answer test set evaluation}.
However, \triviaqa{} improves less, and \abr{em} decreases on \squad{}
with augmented training.
The next section inspects examples to understand why augmented training 
accuracy differs on these datasets.

\paragraph{Freebase vs Wikipedia Aliases.}
We present the comparison of using Wikipedia entities and Freebase entities for augmented evaluation and training on TriviaQA.
We show the augmented evaluation and training results in Table~\ref{tab:trivia_results}. Using Wikipedia entities increases in EM score under augmented evaluation (e.g., the baseline model scores 54.13 under Wiki-expanded augmented evaluation, as compared to 51.75 under Freebase-expanded augmented evaluation). This is mainly because TriviaQA answers have more matches in Wikipedia titles than in Freebase entities. 
On the other hand, the difference between the two alias sources is rather small for augmented training. For example, using Wikipedia for answer expansion improves the baseline from 49.31 to 49.42 under single-answer evaluation, while using Freebase improves it to 49.53.

%


%
%
%

\section{Analysis: Does \abr{qa} Retain that Dear Perfection with
  another Name?}
\label{sec:analysis}

A sceptical reader would rightly suspect that accuracy is only going
up because we have added more correct answers.
Clearly this can go too far\dots if we enumerate all finite length
strings we could get perfect accuracy.
This section addresses this criticism by examining whether the new
answers found with augmented training and evaluation would still
satisfy user information-seeking needs~\cite{voorhees2019evolution} for both the training and test sets.

\begin{table}[t]
\small
\center
\begin{tabular}{lccc}
\toprule
 & {\bf \abr{nq}} & {\bf \squad{}} & {\bf \triviaqa{}}   \\
 \midrule
Correct & 48 & 31 & 41  \\
Debatable & 0 & 2 & 3  \\
Wrong & 1 & 16 & 6  \\
Invalid & 1 & 1 & 0\\
\midrule 
Non-equivalent & 1 & 5 & 2  \\
Wrong context & 0 & 1 & 1  \\
Wrong alias & 0 & 10 & 3  \\
 \bottomrule
\end{tabular}
\caption{Annotation of fifty sampled augmented training examples from
  each dataset.  Most training examples are still correct
  except for \squad{}, where additional answers are incorrect a third
  of the time.  How the new answers are wrong is broken down in the
  bottom half of the table.}
\label{tab:train_annotation}
\end{table}


\begin{table}[t]
\small
\center
\begin{tabular}{lccc}
\toprule
 & {\bf \abr{nq}} & {\bf \squad{}} & {\bf \triviaqa{}}  \\
 \midrule
Correct & 48 & 47 & 50  \\
Wrong & 1 & 1 & 0  \\
Debatable & 1 & 1 & 0 \\
Invalid & 0 & 1 & 0 \\
 \bottomrule
\end{tabular}
\caption{Annotation of fifty test questions that went from incorrect
  to correct under augmented evaluation.  
  Most changes of correctness are deemed valid by human annotators across all three datasets.
  }
\label{tab:test_annotation}
\end{table}



\begin{table}[tb]
    \centering
  
    \small
    \footnotesize

   \begin{tabular}{p{7cm}}
        \toprule
     \rowcolor{gray!50}
             \textbf{Q}: What city in France did the torch relay start at?
        \textbf{P}: Title: 1948 summer olympics. The torch relay then
     run through Switzerland and \ul{France} \dots         \\       
        \textbf{A}: Paris \\
        \textbf{Alias:} France \\
        \textbf{Error Type}: Non-equivalent Entity \\

     \rowcolor{gray!50}
        \textbf{Q}: How many previously-separate phyla did the 2007 study reclassify?     \\
        \textbf{P}: Title: celastrales. In the APG \ul{III} system,
     the celastraceae family was expanded to consist of these five
     groups \dots \\
        \textbf{A}: 3 \\
        \textbf{Alias}: III \\
        \textbf{Error Type}: Wrong Context \\

     \rowcolor{gray!50}
        \textbf{Q}: What is Everton football club's semi-official club nickname?\\     
        \textbf{P}: Title: history of \ul{Everton F. C.}  Everton
     football club have a long and detailed history \dots      \\
        \textbf{A}: the people's club \\
        \textbf{Alias}: Everton F. C. \\
        \textbf{Error Type}: Wrong Alias
        \\


        


     \bottomrule
        \caption{How adding equivalent answers can go wrong.  While
     errors are rare (Table~\ref{tab:train_annotation} and~\ref{tab:test_annotation}), these errors are representatives of mistakes. 
     The examples are taken from \squad{}.}
    \label{tb:case}
  
    \end{tabular}
  \end{table}


\paragraph{Accuracy of Augmented Training Set.}

We annotate fifty passages that originally lack an answer but do have
an answer from the augmented answer set
(Table~\ref{tab:train_annotation}).
We classify them into four catrgories: correct, debatable, and wrong answers, 
as well as invalid questions that are ill-formed or unanswerable
due to annotation error.
The augmented examples are mostly correct for \abr{nq}, 
consistent with its \abr{em} jump with augmented training.
%
However, augmentation often surfaces wrong augmented answers
for \squad{}, which explains why the \abr{em} score drops with  
augmented training.

We further categorize \emph{why} the augmentation is wrong into three
categories (Table~\ref{tb:case}):
%
(1) Non-equivalent entities, where the underlying knowledge base has a
mistake, which is rare in high quality \abr{kb}s; (2)
Wrong context, where the corresponding context is not answering the question;
(3) Wrong alias, where the question  
asks about specific alternate forms of an entity but the prediction is another alias of the entity.
%
%
%
This is relatively common in \squad{}.  We speculate this is a
side-effect of its creation: users write questions given a Wikipedia
paragraph, and the first paragraph often contains an entity's aliases
(e.g., ``Vladimir Ilyich Ulyanov, better known by his alias Lenin, was
a Russian revolutionary, politician, and political theorist''), which
are easy questions to write.


  \paragraph{Accuracy of Expanded Answer Set.}

Next, we sample fifty \emph{test} examples that models get wrong under
the original evaluation but that are correct under augmented
evaluation.
We classify them into four catrgories: correct, debatable, wrong answers, 
and the rare cases of invalid questions.
Almost all of the examples are indeed correct (Table~\ref{tab:test_annotation}),
demonstrating the high precision of our answer expansion for augmented evaluation.
In rare cases, for example, for the question ``Who sang the song Tell Me Something
Good?'', the model prediction \ul{Rufus} is an alias entity, but the
reference answer is \ul{Rufus and Chaka Khan}.
The authors disagree whether that would meet a user's
information-seeking need because Chaka Khan, the vocalist, was part of the band Rufus.
Hence, it was labeled as debatable.

\section{Related Work: Refuse thy Name}
\label{sec:related}



%

\paragraph{Answer Annotation in \abr{qa} Datasets.} Some \abr{qa} datasets such as \abr{nq} and TyDi~\cite{TyDiQA} $n$-way
annotate dev and test sets where they ask different annotators to
annotate the dev and test set.  However, such annotation is costly and
the coverage is still largely lacking
(\textit{e.g.}, our alias
expansion obtains many more answers than \abr{nq}'s original multi-way
annotation).
\ambigqa{}~\cite{AmbigQA} aims to address the problem of ambiguous
\emph{questions}, where there are multiple interpretations of the
same question and therefore multiple correct answer classes (which
could in turn have many valid aliases for each class).
We provide an orthogonal view as
we are trying to expand equivalent answers to any 
given gold answer while \ambigqa{} aims to 
cover semantically different but valid answers. 


\paragraph{Query Expansion Techniques.} Automatic query expansion has been used to improve information retrieval~\cite{QEsurvey}. 
Recently, query expansion has been used in \abr{nlp} applications such as document re-ranking~\cite{BERT-QE} and passage retrieval 
in \odqa{}~\cite{Qi2019AnsweringCO,GAR}, with the goal of increasing accuracy or recall. 
Unlike this work, our answer expansion aims to improve \emph{evaluation} of \qa{} models.

\paragraph{Evaluation of QA Models.} There are other attempts to
improve \qa{} evaluation. 
\citet{EvalQA} find that current automatic metrics do not correlate well with human judgements, which motivated
\citet{MOCHA} to construct a dataset with human annotated scores of
candidate answers and use it to train a BERT-based regression model as
the scorer.
\citet{feng-19} argue for instead of evaluating \abr{qa} systems
directly, we should instead evaluate downstream \emph{human} accuracy
when using \qa{} output.
Alternatively, \citet{Risch2021SemanticAS} use a cross-encoder to measure the semantic similarity between predictions and gold answers. 
For the visual \qa{} task, \citet{VQA-alias} incorporate alias answers in visual \qa{} evaluation. 
In this work, instead of proposing new evaluation metrics, 
we improve the evaluation of \odqa{} models by augmenting 
gold answers with alias from knowledge bases.
\section{Conclusion: Wherefore art thou Single Answer?}
\label{sec:conclusion}

Our approach for matching entities in a \abr{kb} is a simple approach
to improve \abr{qa} accuracy.
We expect future improvements---\textit{e.g.,}, entity linking 
source passages would likely improve precision at the cost of recall.
Future work should also investigate the role of context in deciding the 
correctness of predicted answers.
%
Beyond entities, future work should also consider other types of answers 
such as non-entity phrases and free-form expressions.

As the \abr{qa} community moves to \abr{odqa} and multilingual
\abr{qa}, robust approaches will need to holistically account for
unexpected but valid answers.
This will better help users, use training data more efficiently, and
fairly compare models.
\section*{Acknowledgements}

We thank members of the UMD CLIP lab, the anonymous reviewers and
meta-reviewer for their suggestions and comments.
Zhao is supported by the Office of the Director of National Intelligence (\abr{odni}), Intelligence Advanced
Research Projects Activity (\abr{iarpa}), via the \abr{better} Program contract 2019-19051600005. 
Boyd-Graber is supported by \abr{nsf} Grant \abr{iis}-1822494.
Any opinions, findings, conclusions, or recommendations expressed here are those of the authors and
do not necessarily reflect the view of the sponsors.

\bibliographystyle{style/acl_natbib}
\bibliography{bib/journal-full,bib/chenglei,bib/jbg}


\end{document}